\documentclass[table,dvipsnames,twoside,runningheads]{llncs}

\usepackage{hyperref}

\usepackage{breakcites}

\usepackage{amsmath}
\usepackage{amssymb}
\usepackage[capitalise]{cleveref}
\usepackage{algorithm}
\usepackage{algorithmicx}
\usepackage[compatible]{algpseudocode}
\usepackage{newfloat}
\usepackage{etoolbox}
\usepackage{multirow}
\usepackage[font={footnotesize}]{caption}

\usepackage{subfigure} 

\usepackage{enumitem}
\usepackage{mathtools}

\usepackage{xspace}

\newcommand{\methodlong}{WalkingTime: Dynamic Graph Embedding Using Temporal-Topological Flows\xspace}

\usepackage{graphicx}
\graphicspath{{figures/}{plotting/FIGS/}}

\usepackage{xcolor}

\newcommand{\COMMENTc}[1]{{\small\color{OliveGreen}\COMMENT{#1}}}

\usepackage{framed}

\begin{document}
\setlength{\abovedisplayskip}{0pt}
\setlength{\belowdisplayskip}{0pt}
\setlength{\abovedisplayshortskip}{0pt}
\setlength{\abovedisplayshortskip}{0pt}

\title{\methodlong}
\titlerunning{WalkingTime}

\author{David Bayani\orcidID{0000-0001-5811-6792}} 
\authorrunning{David Bayani}

\institute{Computer Science Department\\Carnegie Mellon University, Pittsburgh, PA 15213, USA\\
\email{dcbayani@alumni.cmu.edu}
}

\maketitle

\setlist{nolistsep}

\begin{abstract}
Increased attention has been paid over the last four years to
dynamic network embedding.
Existing dynamic embedding methods, however, consider the problem
as limited to the evolution of a topology over a sequence of global, discrete
 states. We propose a novel embedding algorithm, WalkingTime, based on a 
fundamentally different
handling of time, 
allowing for the local consideration of continuously occurring phenomena; while
others consider global time-steps to be first-order citizens of the dynamic
environment, we hold flows comprised of temporally and topologically local interactions as our primitives,
without any discretization or alignment of time-related attributes being necessary.\footnote{We
 note that the vast majority of the work and writing presented in this paper
 was done in September to November of 2018 (see, for instance, the hashes provided at \cite{bayani_david_2021_5548589}). For the most part, recent modifications 
to this writeup 
 only streamlined
 portions (e.g., cut content) in order to enable more immediate presentation. Some visuals were mildly improved.}

\end{abstract}

\section{Introduction and Related Work}

Graph embeddings are a collection of techniques for converting aspects of a network
into a vector based (at least partially) on network topology, typically considered
as mapping the high-dimensional structure of the network itself
into a significantly lower dimensional space. This broad endeavor most commonly
is concerned with representation of individual nodes
\cite{hamilton2017representation,cai2018comprehensive}.

The embedding task is often performed by using matrix factorization methods and
neural-network based approaches. 
In addition to diversity resulting from problem-specific modifications,
embedding techniques also vary in 
regard to the balance they strike between
reflecting the structures of community membership and reflecting structural equivalence in the original network,
\footnote{For instance, in a three-node chain, proximity-based 
mappings would put maximum distance between the end nodes' embeddings, whereas methods emphasizing
structure would minimize said distance.}, as well as in regards to what order of proximity they consider
(e.g., the value for K in $A^K$ where A is the adjacency matrix).

Neural network based approaches have seen an increase of popularity over the last several
years, spurred in part by their demonstrations  of superior performance in numerous areas.
 These
 approaches can generally be divided into random-walk-based and 
non-random-walk based methods. The former saw recent popularity ushered in by
DeepWalk \cite{Perozzi:2014:DOL:2623330.2623732} and then 
node2vec \cite{node2vec-kdd2016}; both methods borrow advances in 
language modeling, generating an embedding for a given node by feeding fixed-length
random walks starting from that node into the Skip-Gram word embedding model \cite{mikolov2013efficient}, using the paths
as context windows surrounding the specified start node. Unlike
DeepWalk, node2vec allows for biased random walks controlled
by interpretable, user defined parameters, and the two methods differ in how they
perform the non-convex optimization required by Skip-grams.\footnote{DeepWalk uses a hierarchical
soft max, while node2vec uses negative sampling}. 
Of particular note are these methods' scalablity (being trivially parallelizable and memory-efficient), which
allows for considerations of higher-order proximity as desired, and
--- in the case of node2vec and its derivatives --- allows for shifting emphasis between
reflecting community membership 
and structural equivalence \cite{node2vec-kdd2016}.

Among the notable extensions of DeepWalk and node2vec,
HARP is a  meta-scheme that augments sophisticated embedding techniques 
 to intelligently  avoid the many local-optima
common in non-convex objectives, 
doing so by providing initial conditions that are more favorable 
for the base method.
HARP
constructs a multi-granular hierarchy
of networks by carefully merging nodes going from one level to the next, and then produces
embeddings for each level going from most-granular to least-granular (i.e., from the least detailed network to the most detailed), using the 
final embedding from the previous layer as initial conditions of the next layer \cite{chen2017harp}. 
HARP has been shown to consistently boost performance for a variety of representation-learning techniques \cite{hamilton2017representation}.

Meta-schemes have also been utilized to incorporate attribute information into 
node embeddings, metapath2vec \cite{dong2017metapath2vec} being a chief example.
Working over networks that may have node and edge attributes,
metapath2vec builds off of node2vec by biasing the random walk based on
user-provided templates which specify sequences of attribute-values which
may occur during a random walk. For instance, if $A$,$B$,$C$ and $D$ are attribute
values, a user may require that all random walks consist of paths who sub-sequences
have nodes with attributes
$< A, B, C>$, $<C,D,D>$, or $<D,A>$ ($<\cdots>$ being
sequence constructor notation), in which case the random walk could only visit
a node with attribute $D$ at most twice in a row.

Moving further to enrich the information carried in embeddings,
dynamic graph embedding has received increase attention recently - for 
    instance \cite{Zhu2016TNE7511675,dudynamic,liang2018dynamic,goyal2018dyngem,zhou2018dynamic,yu2018netwalk,li2017attributed,ma2018depthlgp}
     all were published in major venues
    within the last two years,\footnote{The ``two years'' cited here are
    in respect to when this content was originally written; in respect to 
    the time that this write-up has been released on ArXiv, it would be roughly
    five years ago.} while
    \cite{hamilton2017representation,cai2018comprehensive,chen2018tutorial,zhang2018network}, 
    among others, noted the deficit of dynamic graph embedding methods just two
    years ago.\footnote{See prior footnote in regard to the timeline.} 
In this embedding literature, the pervading models of dynamics are global, discrete,
and strictly progress forward in time.
That is, at a time $T + 1$, all areas
of the network in question are updated to include changes since $T$ - effectively
every node receives the compiled changes that have occurred in the time window
$(-\infty , T + 1]$~ \cite{Zhu2016TNE7511675,dudynamic,liang2018dynamic,zhou2018dynamic,goyal2018dyngem,yu2018netwalk,li2017attributed,ma2018depthlgp}.
While these time-steps might have non-uniform window sizes or other sophisticated methods of
selecting analysis times, the point
 remains that the network growth is cut into ``snap-shots'', framing evolution as discrete
and globally-effecting process; this is unnatural given
that in many real systems, events are typically 
a function of local interactions, and many 
distinguishing behaviors may occur over time-scales much shorter than the
collective behavior of the network as it evolves
\cite{leskovec2005graphs}. 
Prior literature  considered a number of
time steps that was order of magnitudes smaller
than either the node set or edge set - for example, in \cite{Zhu2016TNE7511675}, only 13 time steps
are considered in a graph with 200,000 nodes and close to two million edges.\footnote{Their is a question of 
scalability (i.e., time and memory limitations)  and robustness (e.g., not washing-out desired signals) in 
increasing the number of time steps.} 
In addition to loss of fine-grain information, such
an approach requires a global ordering of events relative to each other in order to construct
the sequence of graphs considered. 
Further still, such schemes quite likely privilege relationships between events that happen to be 
co-located in the same discrete step over those that
artificially land across a divide, despite the fact that events close to the separation point
may be temporally closer to one another than to any other members of their respective partitions.
In short, these models of dynamic graphs are, for many dynamic processes,  unnatural and can fundamentally
impact the view which down-stream embedding algorithms have of the network 
dynamics

\begin{figure}[h!]
\includegraphics[width=\linewidth]{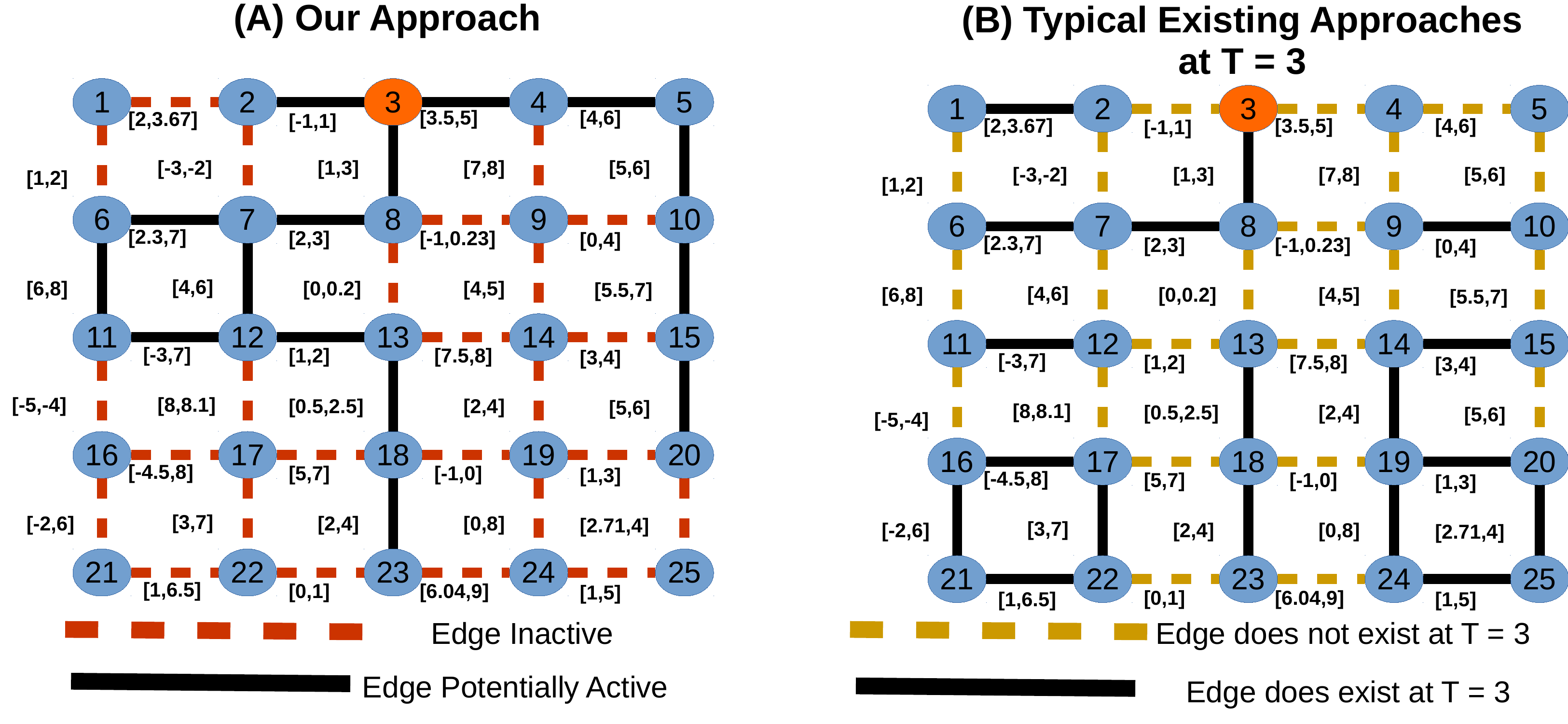}
\caption{
Illustration comparing our approach to typical dynamic graph embedding approaches in an unattributed, unweighted 5-by-5 grid.
In this scenario, we consider embedding the node highlighted in orange, node 3. 
In the figures, the edges that are traversable are shown in black, while edges that exist in the static graph but are not currently considered by the algorithms are dashed and highlighted in orange or yellow.}
\label{fig:exampleGraphComparingOurApproachToOthers}
\end{figure}

\section{Proposed Method: WalkingTime}

Our method, WalkingTime, extends node2Vec to the dynamic setting by 
adopting an event-time multi-graph representation and leveraging
temporal information present on edges to inform the random walk.
In contrast to prior work,
WalkingTime has been 
built off of a dynamic network model that does not require global discretization, 
allowing time to remain a continuous value that can be considered locally between the nodes with which the interaction
occurs. Further, our approach allows much of the temporal information to be used raw, removing several areas where performance-critical
decisions previously had to be made up-front.

\newcommand{\BoI}[0]{\textbf{I}}
We consider as our input a multi-graph 
$G^\BoI_{temporal} = (V, E^\BoI_{intervals} \cup E^\BoI_{time points} \cup E^\BoI_{persist})$
where, letting $\pi$ be the projection operator:
\begin{align*}
\mathbb{R}^2_{intervals} =  \{w \in (\mathbb{R} \cup \{\infty, -\infty\}) \times (\mathbb{R}  \cup \{\infty, -\infty\})| \pi_1(w) < \pi_2(w) \}\\
E^\BoI_{intervals} \subset V \times V \times \mathbb{R}^2_{intervals},~~~~E^\BoI_{time points} \subset V \times V \times \mathbb{R}, ~~~~E^\BoI_{persist} \subset V \times V
\end{align*}
Above, $\mathbb{R}^2_{intervals}$ is the set of valid (including potentially indefinite) time intervals.
For ease of future reference, we will define $H = V \times V \times \mathbb{R}^2_{intervals}$. The semantics of these sets of edges are as follows:
\begin{itemize}
\item{ $E^\BoI_{intervals}$ represents interacts between nodes where beginning and end information are present in the data 
    provided. For instance, a phone call may be recorded in such a fashion.}
\item{$E^\BoI_{time points}$ represents interacts where a single time-point is presented in the data, owing to
    there being only one overt action that occurred, but that was presumably undertaken some period before and which 
    will have impact for some unspecified period after. For example, an email may be recorded this way; while 
    the sent-timestamp of an email is considered a single point in time, there is a latent interval of time stretching from
    when factors lend the author to write
    to when the recipient finishes reading.}
\item{$E^\BoI_{persist}$ denotes those relations that have not observably
    changed over the period of study for which the data was collected. A family
    relationship in a social network is an example,
    which may predate and extend beyond the time span studied.}
\end{itemize}

\noindent We transform this input graph into a uniform representation:
\begin{align*}
G_{temporal} = (V, E_{temporal}) ~~~\text{where}~~~E_{temporal} \subseteq H
\end{align*}
Given a user-defined parameter, $\lambda \in \mathbb{R}^{\ge 0}$ --- which we call the ``symmetric window-extension'' --- 
the transformation is done as follows:
\begin{align*}
	E&_{temporal} = \Big\{e \in H | \Big( \exists e' \in E^\BoI_{intervals}. \pi_1(e) = \pi_1(e') \land \pi_2(e) = \pi_2(e') \land \\ 
	& \pi_1(\pi_3(e)) = \pi_1(\pi_3(e')) - \lambda \land \pi_2(\pi_3(e)) = \pi_2(\pi_3(e')) + \lambda\Big) ~ \bold{\lor} ~ \Big( \exists e' \in E^\BoI_{time points}. \\ 
	&\pi_1(e) = \pi_1(e') \land \pi_2(e) = \pi_2(e') \land \pi_1(\pi_3(e)) = \pi_3(e') - \lambda \land \pi_2(\pi_3(e)) =
	 \pi_3(e') + \lambda \Big)~\bold{\lor}\\
	 &  \Big( \exists e' \in E^\BoI_{persist}. \pi_1(e) = \pi_1(e') \land \pi_2(e) = \pi_2(e') \land \pi_3(e) = (-\infty, \infty) \Big) \Big\}
\end{align*}
The parameter $\lambda$ represents the typical latent lag-time between events that cause an interaction - which we presume, here, 
are dependent on previous interactions - and the overt interaction taking place, as well as a post-event window of effect (i.e., triggering another
event to occur). We comment on methods to select this parameter in \cref{sec:discussionAndConclusion}.

As stated, our aim is to produce a random walk (and eventually a series of random walks) of form $<u_i \in V | i \in \mathbb{N} \cup \{0\}. i \le l_{walk}>$,
where, as similarly required by node2Vec, $l_{walk}$ is a user defined non-negative integer specifying the number of steps in the walk.
To produce our random walks using $G_{temporal}$, we introduce one additional structure, a set of active edges at each step in the random walk
, $A_E : \mathbb{N} \cup \{0\} \rightarrow  \mathcal{P}(E_{temporal})$. 
For convenience of notation in the process of defining this structure, we will use the following, given $v, u \in V$ and $E' \subseteq E_{temporal}$:
\vspace{-\topsep}
\begin{framed}
\vspace{-\topsep}
$\begin{aligned}[t]
\vspace{-\topsep}
&E' \! \upharpoonright \! v \! \! \! &\coloneqq &\{e \in E' | \pi_1(e) = v \lor \pi_2(e) = v\} \\
&E' \! \upharpoonright \! (v, u) \! \! \! &\coloneqq &\{e \in E' | \{\pi_1(e),  \pi_2(e)\} = \{v, u\}\}
\vspace{-\topsep}
\end{aligned}$
\vspace{-\topsep}
\end{framed}
\vspace{-\topsep}
\noindent In words, these are the multi-edges present in $E'$ which are incident on the nodes listed in the restriction.
Note that $E' \! \upharpoonright \! (v, v)$ is only non-empty if node $v$ has explicit self-loops in $E'$.

We define the allowable random walks starting from a node $u_0$ recursively:
\vspace{-\topsep}
\begin{framed}
\vspace{-\topsep}
\begin{itemize}[leftmargin=0in]
\item $A_E(0) = E \! \upharpoonright \! u_0 .$
\item{We require, for all $i \in \mathbb{N} \cup \{0\}$, that  $A_E(i) \upharpoonright (u_{i}, u_{i + 1}) \not= \emptyset$. This
    is a restriction on the possible choice of $u_{i + 1}$ taken by the random walk. \footnotemark}
\item{ $\begin{aligned}[t]
    \forall i \in \mathbb{N} \setminus \{0\}.~A_E(i + 1) = & \{ e \in E_{temporal} \! \upharpoonright \! u_{i+1} | \\ 
         & \exists e' \! \! \in A_E(i) \! \upharpoonright \! u_{i+1}. ~\pi_3(e') \cap \pi_3(e) \not= \emptyset \} \end{aligned}$ }
\end{itemize}
\vspace{-\topsep}
\end{framed}
\vspace{-\topsep}
\footnotetext{Note that this requirement
    in not equivalent to $A_E(i) \upharpoonright u_{i + 1} \not= \emptyset$, which would allow revisiting $u_i$ even without
    $u_i$ having self-loops. }
\noindent Here,  $A_E(i)$ may be understood as the set of ``active edges'' on the $i^{th}$ step of the random walk.
This path construction resembles some of the numerous approaches falling under the heading of
 "time-respecting paths" \cite{HOLME201297}. Further, notice that our approach, compared to the
 approaches based on graph snap-shots, alters both the set of nodes reachable during the random walk as well as the
 distance to the reachable nodes. 

As done in node2Vec, we produce biased random walks that rely on parameters $p, q \in \mathbb{R}^{> 0}$ as follows ( ``$A \gets B$'' indicates assigning a value $B$ to variable $A$\footnote{As opposed to edges or paths on the random walk.}):
\begin{framed}
\vspace{-\topsep}
\begin{itemize}[leftmargin=0in]

\item The set of nodes that may be visited at step $i + 1$ (i.e., the candidates for $u_{i + 1}$ ) is 
    $\mathcal{U}_{i +1} = \{v \in V| A_E(i) \upharpoonright (u_{i}, v_{i + 1}) \not= \emptyset\}$.
\item If $u_{i-1} \in \mathcal{U}_{i+1}$, then the probability of selecting $u_{i +1} \gets u_{i-1}$ is $\alpha p^{-1}$.
\item If $w \in  \mathcal{U}_{i+1} \setminus \{u_{i-1}\}$ is such that there exists $e \in E_{temporal}$
    where $\{\pi_1(e), \pi_2(e)\} = \{w, u_{i-1}\}$ and there exist $e' \in A_E(i) \upharpoonright (u_{i}, v_{i + 1})$
    such that $\pi_3(e') \cap \pi_3(e) \not= \emptyset$, then the probability of selecting $u_{i +1} \gets u_{i-1}$ is $\alpha$.
\item For all other members of $\mathcal{U}_{i +1}$, the probability of selection is $\alpha q^{-1}$.
 
\end{itemize}
\vspace{-\topsep}
\end{framed}
\vspace{-\topsep}
\noindent In the above, $\alpha$ is a normalization constant that causes the likelihoods of selection listed above
to be well-defined probabilities (i.e., so that $\sum_{w \in \mathcal{U}_{i+1}}P(u_{i +1} \gets w) = 1$).

\begin{algorithm}
\small
\captionsetup{format=hang}
 \caption{WalkingTime($G^\BoI_{temporal}$, $\lambda$, $l_{walk}$, $n_{walks}$, \\
     \indent $l_{window}$, $n_{optIters}$, $p$, $q$, $d$) }
 \label{algorithm:WalkingTime}
\begin{algorithmic}[1]
\STATE $G_{temporal} \gets$ TransformGraph($G^\BoI_{temporal}$, $\lambda$)
\STATE sampledRandomWalks $\gets$ empty list \COMMENTc{sampledRandomWalks will become a list of sequences}
\FOR{$v \in V$}
   \FOR {$i = 0; i \le n_{walks}; i\! +\! +$}
       \STATE thisWalk $\gets$ temporalRandomWalk($G_{temporal}$, $v$,
                  $l_{walk}$, $p$, $q$) \indent\indent\COMMENTc{Sample a length $l_{walk}$ biased random walk}
       \STATE sampledRandomWalks.append(thisWalk);
   \ENDFOR
\ENDFOR
\STATE{\begingroup\addtolength{\jot}{-0.5em}$\begin{aligned}[t]\text{embeddings} \gets \text{SkipGram}(& \text{sampledRandomWalks}, l_{window}, \\
                                                                         & n_{optIters}, d) \end{aligned}$ \endgroup}
\STATE \textbf{return} embeddings
\end{algorithmic}
 \end{algorithm}

\cref{algorithm:WalkingTime} summarizes how the pieces described above come together.
Of the parameters listed,  $\lambda$ is the
only additional one that WalkingTime uses  that is not found in node2vec.
The remaining parameters, found in the method we build off of, are as follows: 
$d$ is the embedding dimension, $n_{optIters}$ is the number of optimization 
iterations, and parameters $p$ and $q$ ---
introduced by node2vec when innovating over DeepWalk ---
control the emphasis between 
structural and proximal features.

\subsection{Efficient Traversal}

The original node2vec and its reference implementation
called for biased random walks whose 
step-selection distribution 
 required knowing, for each neighbor $v$ of node $u_{t}$, whether $v$ was $u_{t-1}$ or a neighbor of 
$u_{t-1}$.
Due to the potentially large
 neighborhoods and relatively small effective diameters of a
 many real graphs, this computation can be unnecessarily expensive when ultimately a single random 
selection will be made among the neighbors of $u_{t}$. In order to improve the efficiency of the process,
we conduct a form of rejection sampling, considering the neighbors of $u_{t}$ in random order, and stopping
when we select a node based, probabilistically, on its proximity to $u_{t+1}$ when respecting time on the edges.
Specifically, we reinterpret the parameters $p$ and $q$ from node2Vec --- which were originally relative
weight parameters, roughly speaking --- and now take them to directly be probabilities of retention based on a nodes' temporal-topological
proximity to $u_{t-1}$. In the case where the all neighbors are sampled and none have been selected
(as can happen in rejection-sampling), 
alias-sampling is used to ensure a neighbor is drawn with appropriate probability using the information
which, at that point,  
would already be computed for each neighbor.

\subsection{Illustration of Our Approach}
An illustration comparing our approach (A) to 
snap-shot based approaches (B) can be see in \cref{fig:exampleGraphComparingOurApproachToOthers}.
We see that in (A), around node 8, all and only things in times [1,3] may be active, while around node 12, anything from times [-3, 7] may be active; note that, if a time window of [-3,7] or larger was active around node 8, then more edges would be possibly activate than there are - this highlights the non-trivial local nature of our windowing, which other approaches fail to emulate.
Importantly,
{\it notice that not only does (A) have a different reachable set than (B), but the distance and direction of
travel is altered compared to the (simple) static graph}: In order for one of WalkingTime's random walks
to reach node 6 after starting at node 3, it must visit --- in order --- nodes 8, 7, 12, and 11. Notice that
the edge leading from node 6 to node 7 is active only {\it after } node 6 is visited from node 11;
it is possible for a random walk starting at 3 to go from node 7 to node 6  after having visiting nodes
12 and 11 at least once.

In general, it is trivial to show that the length of allowable paths for WalkingTime
are at least as long as the paths present in any time-step in a snapshot-based model, but not
longer than the paths present in a static graph containing one edge to represent
any that would eventually appear.

Examining the illustration for (B), we see that (1) the length and number of 
paths leading from node 3 are shorter and fewer, and (2) that there are seemingly 
many components to the graph. One can regard this second point as a loss of 
granular, local detail; seemingly, for instance, nodes 14, 15, 19, 20, 24, and 25 are 
temporally related, or at least more closely related to each other at this time
than to node 3. Our method (A) suggests, however, that nodes 15 and 20 are 
transitively closer in relation to node 3 than the rest of the aforementioned 
nodes. Further, we see that in (A), node 15 and node 20 are considered 
immediately related, while nodes 14 and 15 are not - this is in contrast to the
alternative, where node 14 and 15 are considered to be in closer proximity than
node 20 is to node 15.

While aggregation across time might alter some of the trends shown here, the 
extensive segregation and lack of fine granularity shown in (B) would still play a 
role, and the nature of the relationships being complied over time would still 
be fundamentally different than those we search for. 
For us in (A), flows are first-order citizens, while
others consider static time-frames to be first-order citizens and flows as derived 
(second-order) objects.

\section{Experiments}

\begin{figure}[h!]
\subfigure[][Initial setup.  ]{
\label{fig:GOLdata_frame0}
\includegraphics[width=0.3\linewidth]{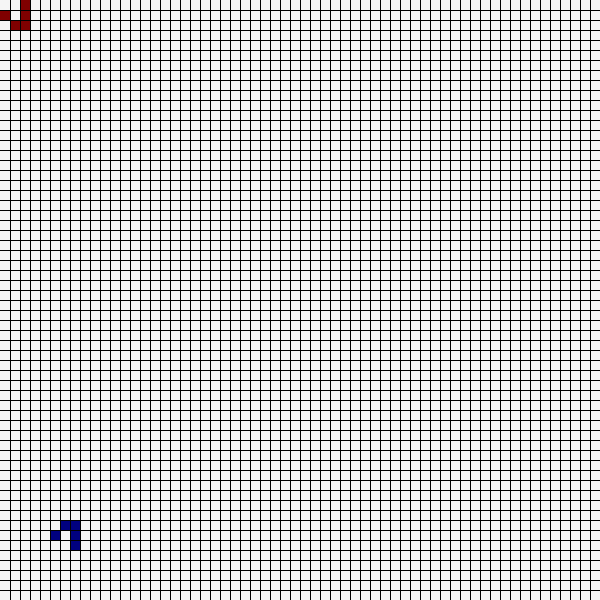}
}
\subfigure[][State and trace of prior positions after 113 steps. 
]{
\label{fig:GOLdata_frame113}
\includegraphics[width=0.3\linewidth]{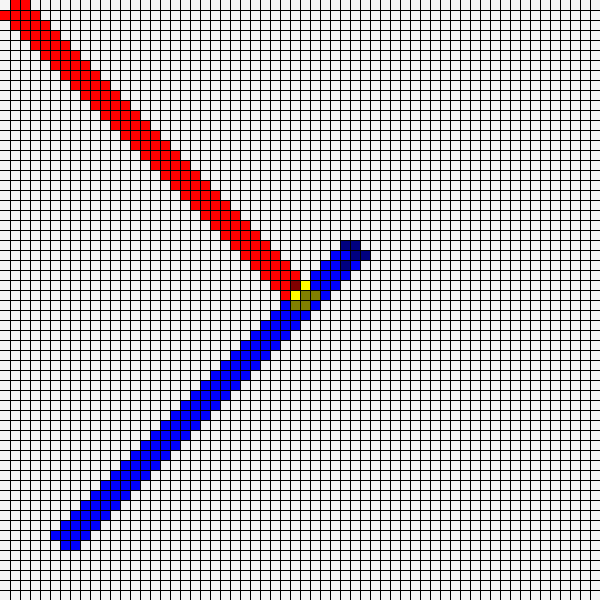}
}
\subfigure[][The final state and full trace of trajectories.  
]{
\label{fig:GOLdata_frame200}
\includegraphics[width=0.3\linewidth]{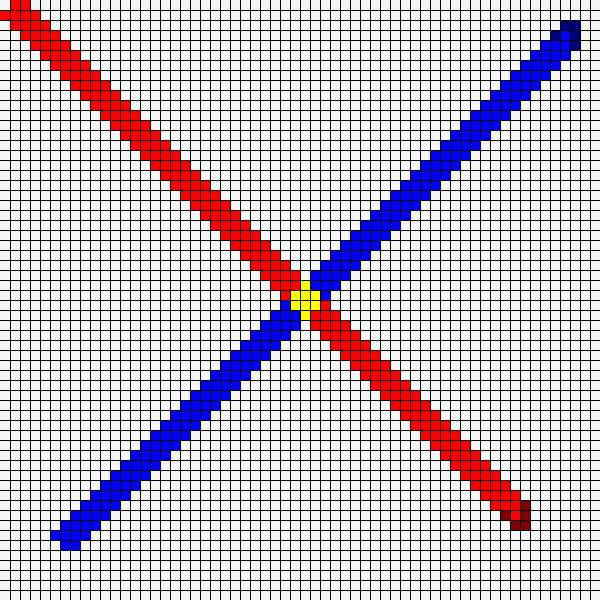}
}
\caption{Figures showing the state of Game of Life at three different time-steps.}
\label{fig:GOLdata} 
\end{figure}

\begin{figure}[h!]
\centering
\includegraphics[width=\linewidth]{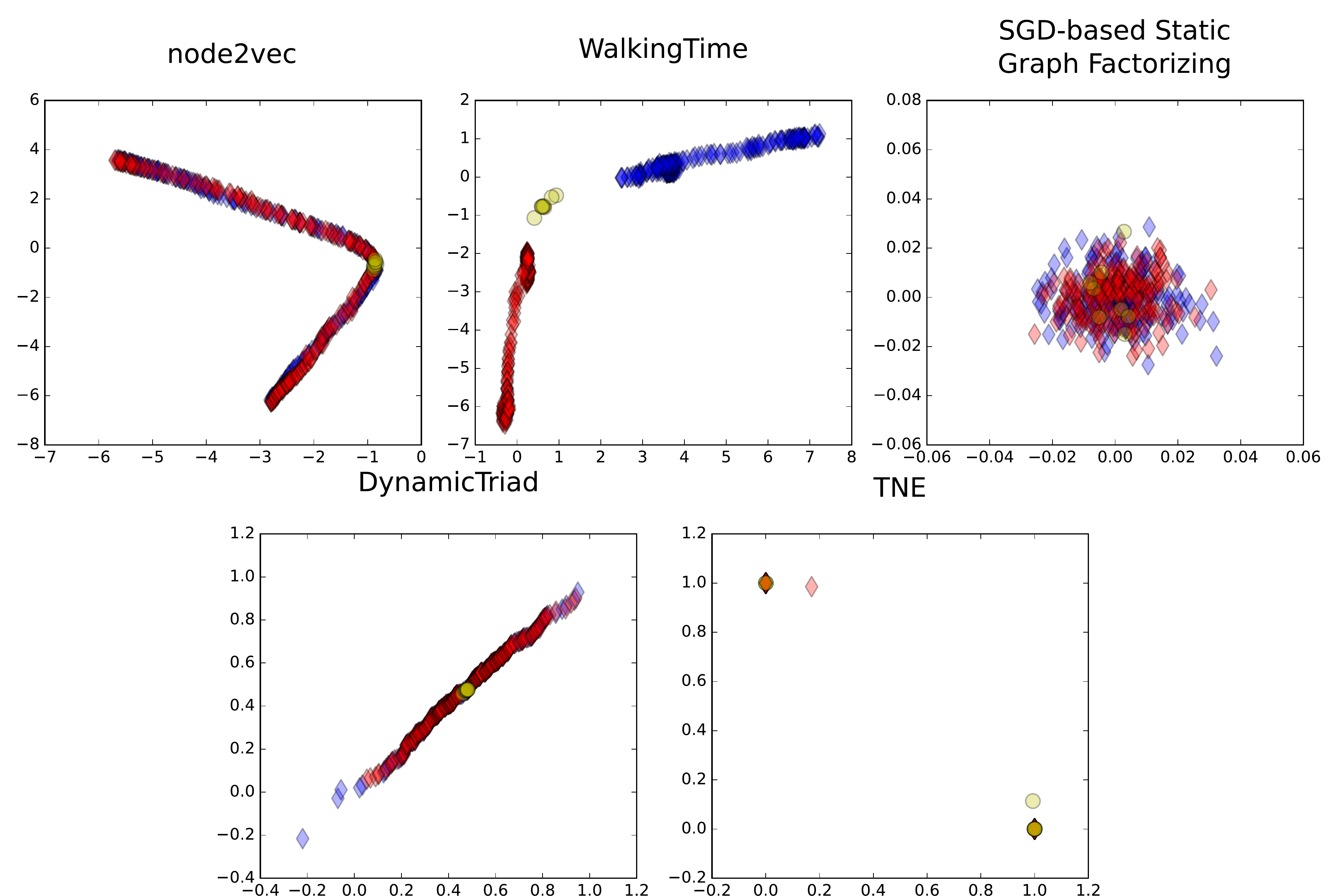}
\caption{Plots of the two-dimensional embeddings produced by various methods on the data generated from \cref{fig:GOLdata}. }
\label{fig:embeddings}
\centering
\end{figure}

	    We briefly demonstrate our approach on a discrete dynamical 
         system that is well formalized, capable of sophisticated behavior, and which is 
         well known to computer scientists: Conway's Game of Life \cite{1970SciAm.223d.120G}. We consider a 
         board with initial configuration shown in \cref{fig:GOLdata_frame0}, where the relative coordinates are (1, 0), (2, 1), (0, 2), (1, 2), and (2, 2) for red, and (53, 5), (52, 6), (54, 7), (53, 7) and (52, 7) for blue. The
colored positions indicated form two ''gliders'', structures which move themselves across the grid as time progresses.
We run the simulation for 200
time-steps. In \cref{fig:GOLdata_frame113}, we see the simulation after 113 steps: the darker cells are active at this step --- note the two gliders still present --- while the lighter blue and red indicate the path taken by the gliders up
to this time (they are not part of the data for this time-step, but shown just to visualize the trajectories). The
yellow is the section where the paths traced by the gliders share a location, although separated in times
when they occupy it. \cref{fig:GOLdata_frame200} shows the states after 200 time-steps, using the same
coloring convention as the previous time-step.

         A temporal graph is produced from this system by assigning one node to each space in the grid that
         at some point was filled, and producing an edge labeled with time $t$ between nodes $v_{i,j}$ and $u_{n,m}$ 
         if and only if $(i,j)$ and $(n,m)$ are neighbors in the grid and are both filled at time $t$. The graph
produced in this fashion
contains 404 nodes and 2200 temporal multi-edges.

We perform a node embedding into two dimensions and plot the results.
	 Our baseline algorithms consist of two methods for static graphs - a scalable graph factorization
based on SGD\cite{Ahmed:2013:DLN:2488388.2488393} (implementation courtesy of \cite{goyal2018Code})
	 and standard node2vec - and
	 two modern methods designed for the temporal setting - TNE\cite{Zhu2016TNE7511675} and DynamicTriad \cite{zhou2018dynamic}.
	TNE performs an adjacency matrix factorization for each graph snap-shot, 
then produces an embedding using a loss that incorporates both reconstruction
        error and cross snap-shot consistency.
	DynamicTriad views network evolution through
         the eyes of triadic closure processes--- taking a local
         understandings of structural evolution, then stitching together
 local views in order to share information globally; the latter feat of global integration
  is accomplished via joint regularization that considers social homophily 
and temporal smoothness.
 WalkingTime and node2vec's parameters were $\lambda=1$,
        $p=10000$, $q=\frac{1}{p}$, and a walk-length of 480. 
All other parameters were the defaults as found on their authors' websites.

Two dimensional embeddings resulting from this process are shown in \cref{fig:embeddings}. The red
and blue represents nodes from the grid that only belonged to one of the trajectories shown in 
\cref{fig:embeddings}, while yellow was assigned to those that occurred in the
cross-over section. As one can see, WalkingTime was the only method that properly sorted the colors.

Further information on experiments can be found in \cref{sec:appendix} and  
\cite{bayani_david_2021_5548589}.

\section{Discussion and Conclusion}
\label{sec:discussionAndConclusion}

We have introduced WalkingTime, a novel temporal graph embedding algorithm that
views dynamic graphs through the lens of temporal-topological flows, as opposed to a 
series of discrete time-steps. Our demonstration shows that this avenue of 
investigation is worth further investigation.

Our approach trivially extends to directed graphs, and can be used in conjunction
 with metapath2vec (\cite{dong2017metapath2vec}) to handle heterogeneous networks.
Further, weighted graphs can be treated by using the weight-distribution in 
a neighborhood to influence the node selection probability, similar to the methods
mentioned in \cite{node2vec-kdd2016} for handling such an extension.

Parameter $\lambda$ may be chosen using a variety of means. A simple approach for
informing the choice of lambda is to produce a histogram approximating the
distribution of time-interval distances among neighboring edges; that is, one may pick 
a random node, $n$, and record the time-difference of two 
randomly select multi-edges incident on it (perhaps
with bias to ensure the multi-edges are between distinct pairs of nodes). If
$[a_1, b_1]$ and $[a_2, b_2]$ are the time-intervals on such a pair of multi-edges (where 
$a_i = b_i$ in the case of representing time-points), then the value
$\frac{1}{2}  ( max(b_1, b_2) - min(a_1, a_2) - (b_1 - a_1) - (b_2 - a_2) )$
may be used in a histogram ;\footnote{The $\frac{1}{2}$ is due to 
$\lambda$ extending both intervals at once.} if this value is zero or negative, the pair of edges already
overlap in time, and if it is positive, the result provides the minimum
$\lambda$ value that would cause the intervals to overlap in time.
While this would provide information about the connectivity of immediate neighborhoods in the
temporal graph, it does not necessarily indicate what impact
$\lambda$ has on the broader graph structure. 
For richer analysis of $\lambda$'s impact on whole-graph statistics,
a variety of sampling based approaches for approximating values of interest
(e.g., effective radius,
connectivity, etc.) may be 
used \cite{hu2013survey,DBLP:conf/sdm/KangTAFL10,DBLP:conf/kdd/LeskovecF06,DBLP:journals/sigkdd/AiroldiC05,DBLP:conf/icassp/KangCF12,DBLP:journals/tkdd/KangTAFL11},
particularly random walks biased toward newly-discovered but not yet visited nodes \cite{DBLP:conf/sigcomm/DoerrB13}.
Alternative to first producing a histogram to then inform a  
human decision for $\lambda$, 
for some measures of interest, it is possible for a user to determine a target value (and tolerance range)
prior to sampling; random walks with mechanisms to vary or sample
$\lambda$ may then more quickly determine a proper parameter choice that results in
a statistic within tolerance of the target value.

Additional material regarding this work can be found at \cite{DBLP:conf/cikm/Bayani20} and \cite{bayani_david_2021_5548589}.
We hope to eventually have the code on GitHub at \url{https://github.com/DBay-ani}.

\bigskip
\noindent\textbf{Acknowledgments.}
We would like to thank Reihaneh Rabbany\orcidID{0000-0003-2348-0353} for her thoughts, guidance, and support while conducting this work.

\bibliographystyle{splncs04}
\bibliography{ref}

\begin{thebibliography}{10}
\providecommand{\url}[1]{\texttt{#1}}
\providecommand{\urlprefix}{URL }
\providecommand{\doi}[1]{https://doi.org/#1}

\bibitem{Ahmed:2013:DLN:2488388.2488393}
Ahmed, A., Shervashidze, N., Narayanamurthy, S., Josifovski, V., Smola, A.J.:
  {Distributed Large-scale Natural Graph Factorization}. In: Proceedings of the
  22nd International Conference on World Wide Web. pp. 37--48. WWW '13, ACM,
  New York, NY, USA (2013). \doi{10.1145/2488388.2488393},
  \url{http://doi.acm.org/10.1145/2488388.2488393}

\bibitem{DBLP:journals/sigkdd/AiroldiC05}
Airoldi, E.M., Carley, K.M.: {Sampling Algorithms for Pure Network Topologies:
  A Study on the Stability and the Separability of Metric Embeddings}. {SIGKDD}
  Explor.  \textbf{7}(2),  13--22 (2005). \doi{10.1145/1117454.1117457},
  \url{https://doi.org/10.1145/1117454.1117457}

\bibitem{DBLP:conf/cikm/Bayani20}
Bayani, D.: {WalkingTime: Dynamic Graph Embedding Using Temporal-Topological
  Flows}. In: Conrad, S., Tiddi, I. (eds.) {Proceedings of the CIKM 2020
  Workshops co-located with 29th ACM International Conference on Information
  and Knowledge Management ({CIKM} 2020), Galway, Ireland, October 19-23,
  2020}. {CEUR} Workshop Proceedings, vol.~2699. CEUR-WS.org (2020),
  \url{http://ceur-ws.org/Vol-2699/invited02.pdf}

\bibitem{bayani_david_2021_5548589}
Bayani, D.: {Material for WalkingTime: Dynamic Graph Embeddings Using
  Temporal-Topological Flows} (Oct 2021). \doi{10.5281/zenodo.5548589},
  \url{https://doi.org/10.5281/zenodo.5548589}

\bibitem{NIPS2001_1961}
Belkin, M., Niyogi, P.: {Laplacian Eigenmaps and Spectral Techniques for
  Embedding and Clustering}. In: Dietterich, T.G., Becker, S., Ghahramani, Z.
  (eds.) Advances in Neural Information Processing Systems 14, pp. 585--591.
  MIT Press (2002),
  \url{http://papers.nips.cc/paper/1961-laplacian-eigenmaps-and-spectral-techniques-for-embedding-and-clustering.pdf}

\bibitem{cai2018comprehensive}
Cai, H., Zheng, V.W., Chang, K.: {A Comprehensive Survey of Graph Embedding:
  Problems, Techniques and Applications}. IEEE Transactions on Knowledge and
  Data Engineering  (2018)

\bibitem{chen2018tutorial}
Chen, H., Perozzi, B., Al-Rfou, R., Skiena, S.: {A Tutorial on Network
  Embeddings}. arXiv preprint arXiv:1808.02590  (2018)

\bibitem{chen2017harp}
Chen, H., Perozzi, B., Hu, Y., Skiena, S.: {HARP: hierarchical representation
  learning for networks}. arXiv preprint arXiv:1706.07845  (2017)

\bibitem{de2013anatomy}
De~Domenico, M., Lima, A., Mougel, P., Musolesi, M.: {The Anatomy of a
  Scientific Rumor}. Scientific reports  \textbf{3}, ~2980 (2013)

\bibitem{DBLP:conf/sigcomm/DoerrB13}
Doerr, C., Blenn, N.: {Metric Convergence in Social Network Sampling}. In: Hui,
  P., Miluzzo, E., Haddadi, H. (eds.) Proceedings of the 5th {ACM} workshop on
  Hot topics in planet-scale measurement, HotPlanet@SIGCOMM 2013, Hong Kong,
  China, August 12-16, 2013. pp. 45--50. {ACM} (2013).
  \doi{10.1145/2491159.2491168}, \url{https://doi.org/10.1145/2491159.2491168}

\bibitem{dong2017metapath2vec}
Dong, Y., Chawla, N.V., Swami, A.: {metapath2vec: Scalable representation
  learning for heterogeneous networks}. In: Proceedings of the 23rd ACM SIGKDD
  International Conference on Knowledge Discovery and Data Mining. pp.
  135--144. ACM (2017)

\bibitem{dudynamic}
Du, L., Wang, Y., Song, G., Lu, Z., Wang, J.: {Dynamic Network Embedding : An
  Extended Approach for Skip-gram based Network Embedding}. In: Lang, J. (ed.)
  Proceedings of the Twenty-Seventh International Joint Conference on
  Artificial Intelligence, {IJCAI} 2018, July 13-19, 2018, Stockholm, Sweden.
  pp. 2086--2092. ijcai.org (2018). \doi{10.24963/ijcai.2018/288},
  \url{https://doi.org/10.24963/ijcai.2018/288}

\bibitem{1970SciAm.223d.120G}
{Gardner}, M.: {Mathematical Games}. Scientific American  \textbf{223},
  120--123 (Oct 1970). \doi{10.1038/scientificamerican1070-120}

\bibitem{goyal2018Code}
Goyal, P., Ferrara, E.: {GEM: A Python package for graph embedding methods}.
  Journal of Open Source Software  (2018).
  \doi{https://doi.org/10.21105/joss.00876},
  \url{https://doi.org/10.21105/joss.00876}

\bibitem{goyal2018dyngem}
Goyal, P., Kamra, N., He, X., Liu, Y.: {DynGEM: Deep Embedding Method for
  Dynamic Graphs}. arXiv preprint arXiv:1805.11273  (2018)

\bibitem{node2vec-kdd2016}
Grover, A., Leskovec, J.: {node2vec: Scalable Feature Learning for Networks}.
  In: Proceedings of the 22nd ACM SIGKDD International Conference on Knowledge
  Discovery and Data Mining (2016)

\bibitem{hamilton2017representation}
Hamilton, W.L., Ying, R., Leskovec, J.: {Representation Learning on Graphs:
  Methods and Applications}. arXiv preprint arXiv:1709.05584  (2017)

\bibitem{HOLME201297}
Holme, P., Saramäki, J.: {Temporal Networks}. Physics Reports
  \textbf{519}(3),  97 -- 125 (2012).
  \doi{https://doi.org/10.1016/j.physrep.2012.03.001},
  \url{http://www.sciencedirect.com/science/article/pii/S0370157312000841},
  temporal Networks

\bibitem{hu2013survey}
Hu, P., Lau, W.C.: {A Survey and Taxonomy of Graph Sampling}  (2013)

\bibitem{DBLP:conf/icassp/KangCF12}
Kang, U., Chau, D.H., Faloutsos, C.: {Pegasus: Mining Billion-Scale Graphs in
  the Cloud}. In: 2012 {IEEE} International Conference on Acoustics, Speech and
  Signal Processing, {ICASSP} 2012, Kyoto, Japan, March 25-30, 2012. pp.
  5341--5344. {IEEE} (2012). \doi{10.1109/ICASSP.2012.6289127},
  \url{https://doi.org/10.1109/ICASSP.2012.6289127}

\bibitem{DBLP:conf/sdm/KangTAFL10}
Kang, U., Tsourakakis, C.E., Appel, A.P., Faloutsos, C., Leskovec, J.: {Radius
  Plots for Mining Tera-byte Scale Graphs: Algorithms, Patterns, and
  Observations}. In: Proceedings of the {SIAM} International Conference on Data
  Mining, {SDM} 2010, April 29 - May 1, 2010, Columbus, Ohio, {USA}. pp.
  548--558. {SIAM} (2010). \doi{10.1137/1.9781611972801.48},
  \url{https://doi.org/10.1137/1.9781611972801.48}

\bibitem{DBLP:journals/tkdd/KangTAFL11}
Kang, U., Tsourakakis, C.E., Appel, A.P., Faloutsos, C., Leskovec, J.: {{HADI:}
  Mining Radii of Large Graphs}. {ACM} Trans. Knowl. Discov. Data
  \textbf{5}(2),  8:1--8:24 (2011). \doi{10.1145/1921632.1921634},
  \url{https://doi.org/10.1145/1921632.1921634}

\bibitem{DBLP:conf/kdd/LeskovecF06}
Leskovec, J., Faloutsos, C.: {Sampling from Large Graphs}. In: Eliassi{-}Rad,
  T., Ungar, L.H., Craven, M., Gunopulos, D. (eds.) Proceedings of the Twelfth
  {ACM} {SIGKDD} International Conference on Knowledge Discovery and Data
  Mining, Philadelphia, PA, USA, August 20-23, 2006. pp. 631--636. {ACM}
  (2006). \doi{10.1145/1150402.1150479},
  \url{https://doi.org/10.1145/1150402.1150479}

\bibitem{leskovec2005graphs}
Leskovec, J., Kleinberg, J., Faloutsos, C.: Graphs over time: densification
  laws, shrinking diameters and possible explanations. In: Proceedings of the
  eleventh ACM SIGKDD international conference on Knowledge discovery in data
  mining. pp. 177--187 (2005)

\bibitem{snapnets}
Leskovec, J., Krevl, A.: {SNAP Datasets: Stanford Large Network Dataset
  Collection}. \url{http://snap.stanford.edu/data} (Jun 2014)

\bibitem{li2017attributed}
Li, J., Dani, H., Hu, X., Tang, J., Chang, Y., Liu, H.: {Attributed Network
  Embedding for Learning in a Dynamic Environment}. In: Proceedings of the 2017
  ACM on Conference on Information and Knowledge Management. pp. 387--396. ACM
  (2017)

\bibitem{liang2018dynamic}
Liang, S., Zhang, X., Ren, Z., Kanoulas, E.: {Dynamic Embeddings for User
  Profiling in Twitter}. In: Proceedings of the 24th ACM SIGKDD International
  Conference on Knowledge Discovery \& Data Mining. pp. 1764--1773. ACM (2018)

\bibitem{ma2018depthlgp}
Ma, J., Cui, P., Zhu, W.: {DepthLGP: Learning Embeddings of Out-of-Sample Nodes
  in Dynamic Networks}. AAAI (2018)

\bibitem{mikolov2013efficient}
Mikolov, T., Chen, K., Corrado, G., Dean, J.: {Efficient Estimation of Word
  Representations in Vector Space}. arXiv preprint arXiv:1301.3781  (2013)

\bibitem{Ou:2016:ATP:2939672.2939751}
Ou, M., Cui, P., Pei, J., Zhang, Z., Zhu, W.: {Asymmetric Transitivity
  Preserving Graph Embedding}. In: Proceedings of the 22Nd ACM SIGKDD
  International Conference on Knowledge Discovery and Data Mining. pp.
  1105--1114. KDD '16, ACM, New York, NY, USA (2016).
  \doi{10.1145/2939672.2939751},
  \url{http://doi.acm.org/10.1145/2939672.2939751}

\bibitem{Perozzi:2014:DOL:2623330.2623732}
Perozzi, B., Al-Rfou, R., Skiena, S.: {DeepWalk: Online Learning of Social
  Representations}. In: Proceedings of the 20th ACM SIGKDD International
  Conference on Knowledge Discovery and Data Mining. pp. 701--710. KDD '14,
  ACM, New York, NY, USA (2014). \doi{10.1145/2623330.2623732},
  \url{http://doi.acm.org/10.1145/2623330.2623732}

\bibitem{Wang:2016:SDN:2939672.2939753}
Wang, D., Cui, P., Zhu, W.: {Structural Deep Network Embedding}. In:
  Proceedings of the 22Nd ACM SIGKDD International Conference on Knowledge
  Discovery and Data Mining. pp. 1225--1234. KDD '16, ACM, New York, NY, USA
  (2016). \doi{10.1145/2939672.2939753},
  \url{http://doi.acm.org/10.1145/2939672.2939753}

\bibitem{yu2018netwalk}
Yu, W., Cheng, W., Aggarwal, C.C., Zhang, K., Chen, H., Wang, W.: {NetWalk: A
  Flexible Deep Embedding Approach for Anomaly Detection in Dynamic Networks}.
  In: Proceedings of the 24th ACM SIGKDD International Conference on Knowledge
  Discovery \& Data Mining. pp. 2672--2681. ACM (2018)

\bibitem{zhang2018network}
Zhang, D., Yin, J., Zhu, X., Zhang, C.: {Network Representation Learning: A
  Survey}. IEEE Transactions on Big Data  (2018)

\bibitem{zhou2018dynamic}
{Zhou}, L., {Yang}, Y., {Ren}, X., {Wu}, F., {Zhuang}, Y.: {Dynamic Network
  Embedding by Modelling Triadic Closure Process}. In: AAAI (2018)

\bibitem{Zhu2016TNE7511675}
Zhu, L., Guo, D., Yin, J., Steeg, G.V., Galstyan, A.: {Scalable Temporal Latent
  Space Inference for Link Prediction in Dynamic Social Networks}. IEEE
  Transactions on Knowledge and Data Engineering  \textbf{28}(10),  2765--2777
  (Oct 2016). \doi{10.1109/TKDE.2016.2591009}

\end{thebibliography}

\appendix
\section{Appendix}
\label{sec:appendix}

{\color{gray} ---Below is a portion of the writing that was under construction for the description of the more comprehensive
experiments. We leave a subset of a content outline and a few notes on what existed, but do not 
give the full details for now. We may return at a later time to provide this information.---}

\subsection{Baseline Algorithms}
\setlist[description]{leftmargin=\parindent,labelindent=\parindent}
\textbf{Static Embeddings}\footnote{The implementations of our static baselines, with the exception of node2vec, come
        courtesy of \cite{goyal2018Code}.}
\itemsep2pt
\begin{description}
    \item [SDNE (\cite{Wang:2016:SDN:2939672.2939753}):] 
    \item [Laplacian Eigenmaps (LAP) (\cite{NIPS2001_1961}):]
    \item [Graph Factorization (\cite{Ahmed:2013:DLN:2488388.2488393}):]
    \item [HOPE (\cite{Ou:2016:ATP:2939672.2939751}):]
    \item [node2vec (\cite{node2vec-kdd2016}):] 
\end{description}
\textbf{Dynamic Embeddings}\itemsep2pt
\begin{description}
    \item [TNE (\cite{Zhu2016TNE7511675}):]
    \item [DynamicTriad (\cite{zhou2018dynamic}):]
\end{description}

\subsection{Evaluation Tasks}

\noindent\underline{\textbf{Node Classification:}} A standard task for node embeddings, we feed the embeddings
 produced by our method and the baselines into classifiers as part of a node-classification task.
 We choose relatively simple classifiers to ensure that the performance reflects qualities of the
 embeddings and their ability to make desired structure apparent, as opposed to agility of a
 sophisticated classifier to pick up on latent signals that might be in otherwise confused / noisy
 data. In order to perform due diligence in ensuring the performance results are not simply artifacts from the
 specifics of one classifier, we use a variety of standard classifiers- KNN, Gaussian Naive Bayes,
 Logistic Regression, and an SVM \textit{\color{gray} [rest redacted for now]}

\noindent\underline{\textbf{Two-Dimensional Node Embedding Plots:}} 

\noindent\underline{\textbf{Latent Graph Reconstruction:}} Given a graph $G_{observed} =$ 
$ (V, E_{observed})$ to produce
embeddings on, we evaluate how the local rankings of distances between node embeddings resemble relations
in a latent graph, $G_{latent} = (V, E_{latent})$, which influences the dynamics. Specifically, for a node $v$
 in $V$ and $k$-dimensional embedding mapping $\phi$, we rank the pairwise distances between $\phi(v)$ and members of 
$\{y \in \mathbb{R}^{k} | \exists u \in V \setminus \{v\}. \phi(u) = y\}$, and produce ROCs /  AUCs to determine
how well this ranking reflects the neighbor relationship in $E_{latent}$. 
Note that this reconstruction scheme is locally linear, not globally - this more appropriately handles
potentially nonlinear scaling between areas of the embedding space than considering a single global
distance cutoff below which edges are considered to exist. 
To avoid computation costs quadratic in the size of the graph's nodes, we randomly sample 1000 nodes from $V$ to do
 full pair-wise comparisons against the rest of the set's members. \textit{\color{gray} [rest redacted for now]} 
\\
\noindent\underline{\textbf{Link Forecasting:}} Given a splitting time, $t_{split}$, we generate embeddings with all information
available up to and including $t_{split}$, and train classifiers to determine if nodes $v_1$ and $v_2$ 
will have links present
at some time after $t_{split}$.
Under the hypothesis that identifying persistent edges are easier, we make sure in evaluation to highlight
both total prediction performance and performance restricted to edges not present at time $t_{split}$. 
While this
task is similar to traditional link prediction tasks, we emphasize that the desired edges are not the result
of missing or partial topological information in the training set, but from lack of knowledge of how the
topology evolves; that is, at any time for which it is available, the topology is not considered to be broken
or downsampled for the sake of evaluation, in contrast to \textit{\color{gray} [rest redacted for now]}

\subsection{Datasets}
\begin{description}
    \item [Simulated Data:] In order to produce synthetic data for initial testing, 
development, and reasonable demonstration, 
we selected a discrete dynamical 
         system that is simple to implement while being capable of sophisticated behavior and which is 
         well known to computer scientists: Conway's Game of Life \cite{1970SciAm.223d.120G}. \textit{\color{gray} [rest redacted for now]}
    \item [HiggsTwitter:\cite{snapnets,de2013anatomy}:]\footnote{\url{https://snap.stanford.edu/data/higgs-twitter.html}}  This
         data was collected from Twitter following the announcement of the Higgs boson's discovery on 
         July 4th, 2012. In total, the data spans seven days: three days prior to the day of the announcement (the $1^{st}$ of July)
         and three days after (the $7^{th}$ of July). We consider two graphs where nodes are accounts: One static
         graph consisting of friendship / social relations between accounts, and one dynamic multi-graph where
         an edge between two nodes represents a retweet, reply, or mention, with a label indicating the
         time of such activity. We will refer to the former network as the social graph and the later 
         as the activity graph. \textit{\color{gray} [rest redacted for now]}
    \item [DBLP:] 
    \item [CSTraces Weather:] \url{http://skuld.cs.umass.edu/traces/sensors/weather.tar.gz} \textit{\color{gray} [rest redacted for now]}
    \item [Digg:]
    \item [blogCatalog:]
\end{description}

\end{document}